\documentclass[runningheads,a4paper]{llncs}
\usepackage{graphicx}
\usepackage{threeparttable}
\usepackage{multirow}
\usepackage{caption}
\usepackage{colortbl}

\begin{document}
\captionsetup{font={scriptsize}}
\captionsetup[table]{aboveskip=1pt,belowskip=-20pt}
\captionsetup[figure]{aboveskip=1pt,belowskip=-10pt}
\mainmatter  

\title{Gland Instance Segmentation by Deep Multichannel Side Supervision}

\titlerunning{Gland Instance Segmentation by Deep Multichannel Side Supervision}

%
%


%
%
\author{Yan Xu\inst{1,2} \and Yang Li\inst{1} \and
Mingyuan Liu\inst{1} \and Yipei Wang\inst{1} \and Maode Lai\inst{3} \and \\ Eric I-Chao Chang\inst{2}\thanks{Corresponding author. Email: echang@microsoft.com}}

\institute{State Key Laboratory of Software Development Environment and Key Laboratory of Biomechanics and Mechanobiology of Ministry of Education and Research Institute of Beihang University in Shenzhen, Beihang University, Beijing 100191, China
\and
Microsoft Research, Beijing 100080, China
\and
Department of Pathology, School of Medicine, Zhejiang University, China
}

\authorrunning{Yan Xu et al.}

\maketitle

\begin{abstract}
In this paper, we propose a new image instance segmentation method that segments individual glands (instances) in colon histology images.
This is a task called instance segmentation that has recently become increasingly important.
The problem is challenging since not only do the glands need to be segmented from the complex background, they are also required to be individually identified.
Here we leverage the idea of image-to-image prediction in recent deep learning by building a framework that automatically exploits and fuses complex multichannel information, regional and boundary patterns, with side supervision (deep supervision on side responses) in gland histology images. 
Our proposed system, deep multichannel side supervision (DMCS), alleviates heavy feature design due to the use of convolutional neural networks guided by side supervision.
Compared to methods reported in the 2015 MICCAI Gland Segmentation Challenge, we observe state-of-the-art results based on a number of evaluation metrics.

\end{abstract}

\section{Introduction}
\label{sec:intro}  
Recent progress in deep learning technologies has led to explosive development in machine learning and computer vision for building systems that have shown substantial improvements in a wide range of applications such as image classification \cite{krizhevsky2012imagenet,vggnet} and object detection  \cite{girshick2015fast}.
The fully convolutional neural networks (FCN) \cite{long2015fully} enable end-to-end training and testing for image labeling; holistically-nested edge detector (HED) \cite{xie15hed} learns hierarchically embedded multi-scale edge fields to account for the low-, mid-, and high- level information for contours and object boundaries. FCN performs image-to-image training and testing, a factor that has become crucial in attaining a powerful modeling and computational capability of complex natural images and scenes.

FCN family models \cite{long2015fully,xie15hed} are well-suited for image labeling/segmentation in which each pixel is assigned a label from a pre-specified set. However, they can not be directly applied to the problem where individual objects need to be identified. This is a problem called instance segmentation. In image labeling, two different objects are assigned with the same label so long as they belong to the same class; in instance segmentation, objects belonging to the same class also need to be identified individually, in addition to obtaining their class labels.
Recent work developed in computer vision \cite{dai2015instance} shows interesting results for instance segmentation but a system like  \cite{dai2015instance} is for segmenting individual objects in natural scenes. With the proposal of fully convolutional network (FCN) \cite{long2015fully}, the "end-to-end" learning strategy has strongly simplified the training and testing process and achieved state-of-the-art results in solving the segmentation problem back at the time. To refine the partitioning result of FCN, \cite{krahenbuhl2012efficient} and \cite{zheng2015conditional} integrate Conditional Random Fields (CRF) with FCN. However, they are not able to distinguish different objects leading to failure in instance segmentation problem. DCAN \cite{chen2016dcan} and U-net \cite{ronneberger2015u} are two instance aware neural networks based on FCN with reasonable performance.

\begin{figure}[htbp] 
 \centering
 \includegraphics[width=4.1in]{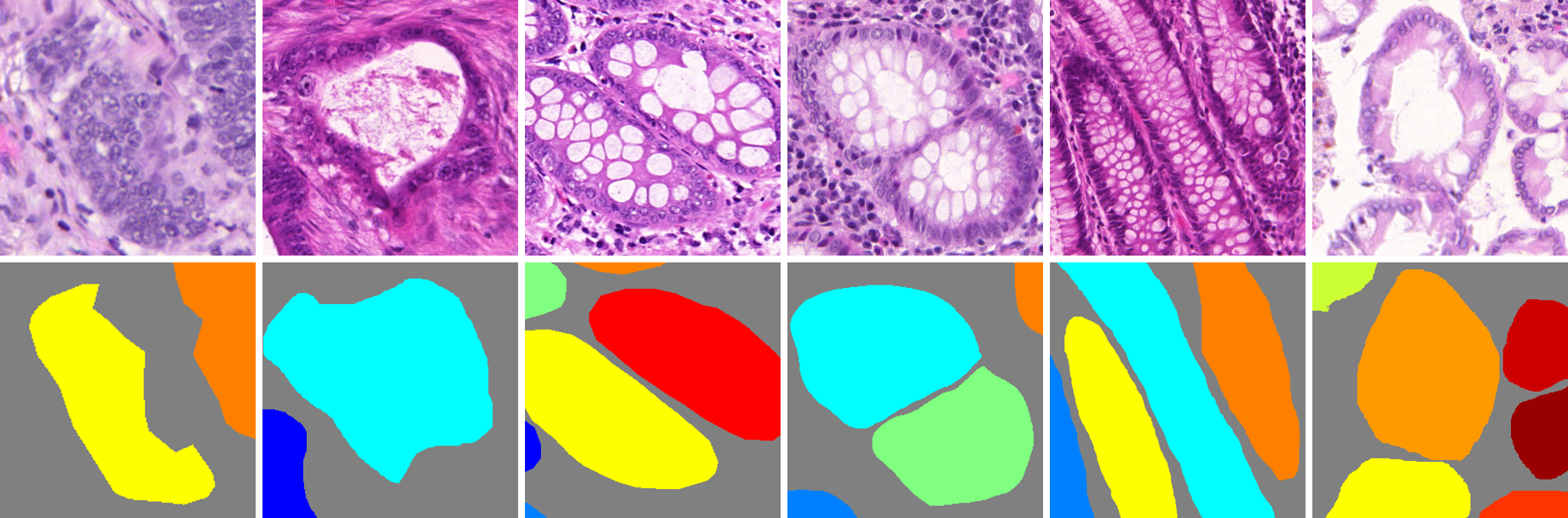} 
 \caption{Gland Haematoxylin and Eosin (H\&E) stained slides and ground truth labels. Images in the first row exemplify different glandular structures. Characteristics such as heterogeneousness and anisochromasia can be observed in the image. The second row shows the ground truth. To achieve better visual effects, each color represents an individual glandular structure.}
 \label{fig.1} 
\end{figure} 

The intrinsic properties of medical image pose plenty of challenges in instance segmentation \cite{dimopoulos2014accurate}. First of all, the objects are in heterogeneous shapes, which make it difficult to use mathematical shape models to achieve the segmentation task. Take colorectal cancer histology image as an example (Fig.1). When the cytoplasm is filled with mucinogen granule the nucleus is extruded into a flat shape whereas the nucleus appears as a round or oval body after secreting. Second, variability of intra- and extra- cellular matrix is often the culprit leading to anisochromasia. Therefore, the background portion of medical images contains more noise like intensity gradients, compared to natural images.

In this paper, we aim to developing a practical system for instance segmentation in gland histology images.
We make use of multichannel learning \cite{szegedy2015going}, region and boundary cues using convolutional neural networks with side supervision, and solve the instance segmentation issue in the gland histology image. Our algorithm is evaluated on the dataset provided by MICCAI 2015 Gland Segmentation Challenge Contest \cite{sirinukunwattana2016gland,sirinukunwattana2015stochastic} and achieves state-of-the-art performance.

\section{Method}
\subsection{HED-Side Convolution (HED-SC)}
The task of pathology image analysis is challenging yet crucial. The booming development of machine learning provides pathology slide image analysis with copious algorithms and tools. Although FCN has been shown to be excellent \cite{long2015fully}, due to the loss of boundary information during downsampling, FCN fails to distinguish instances in certain classes. To conquer this challenge, HED learns rich hierarchical representations under the guidance of deep supervision with each layer capable of carrying out an edge map at a certain scale. Thus the HED model is naturally multi-scale. Combining the side-outputs together, the weighted-fusion layer integrates the features obtained from different levels yielding superior results (for more details on HED, see \cite{xie15hed}). Since our model performs the edge detection on the basis of pixelwise prediction, the transformation from the region feature to boundary feature is required. Hence, the original HED model is modified by adding two convolution layers in each side output path and the HED-SC model is born.
In this paper, we build a multichannel model (Fig.2) that accomplishes the task of instance segmentation in the gland histology image.

\begin{figure}[!h] 
 \centering
 \includegraphics[width=4.1in]{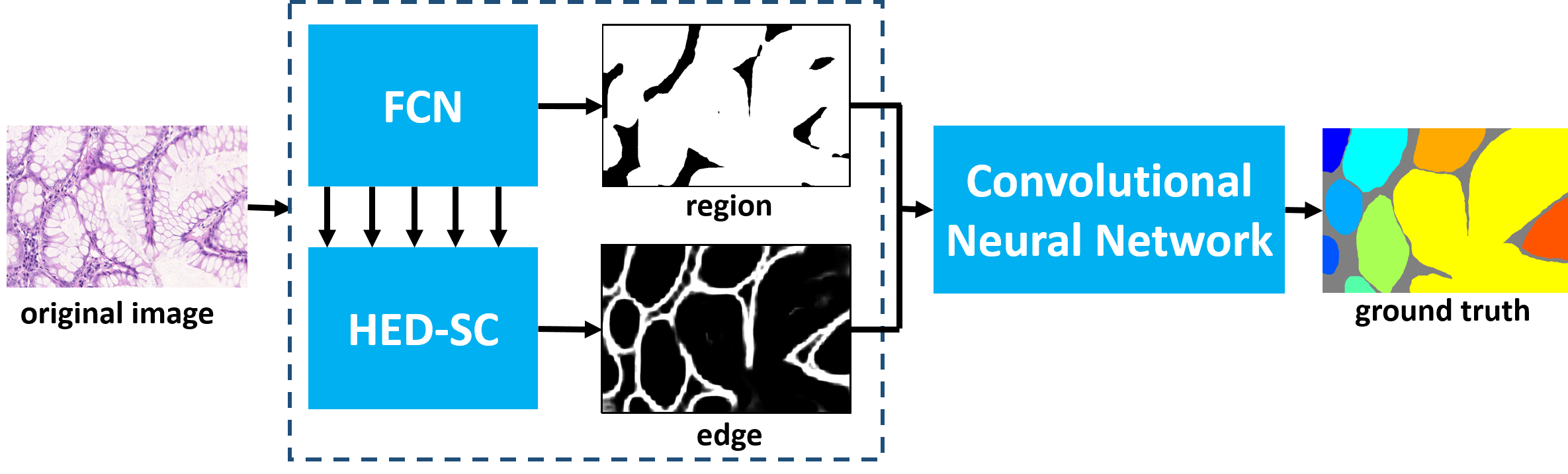} 
 \caption{ Figure above illustrates a brief structure of DMCS. The black arrows represent the forward learning progress. FCN, the region channel, yields the prediction of regional probability maps. HED-SC, the edge channel, outputs the result of boundary detection. A convolution neural network concatenates features generated by different channels and produces segmented instances. } 
\end{figure} 

\begin{figure}[t] 
 \tiny
 \centering
 \includegraphics[width=4.5in]{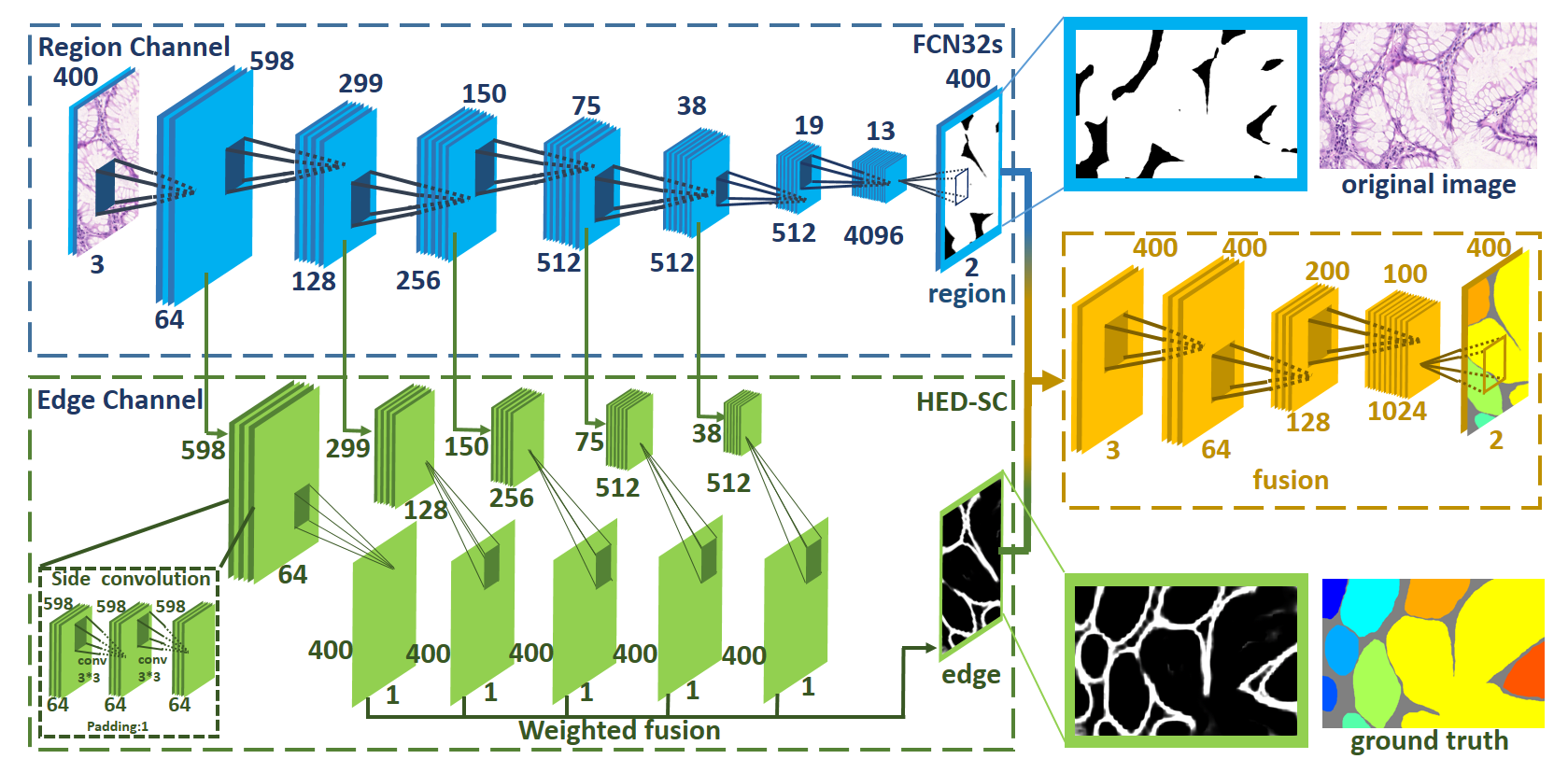} 
 \caption{Illustrates the deep multichannel side supervision model. The region channel engaged in producing a coarse pixel prediction of which the structure is identical to FCN32s [8]. At the first convolutional layer, padding of 100 pixels is involved as Long does [8]. The output of this channel achieved via the strategy of in-network up-sample layers and crop layers is the same size as the input images. Boundary information is obtained by the HED-SC channel of DMCS inspired by HED [7]. In this edge detection model side convolution is inserted before all the pooling layers in the FCN32s. Altogether, there are five side convolutions. Learnable weighting is assigned to five output of deep supervisions to produce the final result. The third part in DMCS aims to do instance segmentation based on information of region and boundary. It concatenates the output of the region channel and the HED-SC channel together. This fully convolutional neural network is utilized to process the segmented images.}
 \label{fig.3} 
\end{figure}

\subsection{Multichannel Learning}
There are N images in the training set that can be divided into K categories. Note that K is the number of object categories. We denote our training set by $S=\{(X_{n},Y_{n},Z_{n}),n=1,2,...,N\}$ where $X_{n}=\{x^{(n)}_{j},j=1,2,...,\left|X_{n}\right|\}$ denotes the original input image, $Y_{n}=\{y^{(n)}_{j},j=1,2,...,\left|Y_{n}\right|\}$, ${y}_{j} \epsilon \{0,1,2,...,K\}$ and $Z_{n}=\{z^{(n)}_{j},j=1,2,...,\left|Z_{n}\right|\}$, ${z}_{j}\epsilon \{0,1\}$ denotes the corresponding ground truth label and binary edge map for image $X_{n}$ respectively. For convenience, $X_{n}$ is simplified as X since all the training images are independent. Our goal is to predict the output set Y from the input image X. By multichannel, we emphasize that we exploit basic cues of segmenting images - region context and edge context - as two channels. 


\textbf{Region feature channel} The region feature channel optimizes the pixel-wise prediction $P_{r}$. We fix the parameter $w_{e}$, $w_{f}$ while learning the parameter $w$, $w_{r}$. The parameters in HED-SC and the parameters before the fully connection layer are represented as $w_{e}$ and $w_{r}$ respectively. Parameters in the fuse stage are denoted as $w_{f}$. Shared with both channels, the weights in FCN before $w_{r}$ are represented as $w$. In this stage, our proposed model follows the architecture of FCN. Fully convolutional networks are trained pixel-to-pixel to achieve image semantic segmentation. Given an input image X, we first predict the pixel-to-pixel label $Y^{*}$ where $\mu_{k}$ denotes the $k^{th}$ class output of softmax function and $h(\cdot)$ calculates the activation of neural network:
\begin{equation}
  {P}_{r}\left(y^{*}_{j}=k \mid X;w,{w}_{r} \right)={\mu}_{k}\left(h\left(X,w,{w}_{r} \right) \right),
\end{equation}
The loss function in this stage is:
\begin{equation}
{L}_{r}\left(Y^{*},X,w,{w}_{r} \right)=\sum_{j=1}^{\left|Y^{*} \right|}{l}_{log}\left({P}_{r}\left(y^{*}_{j}=y_{j}\mid X;w,{w}_{r} \right) \right).
\end{equation}
$l_{log}(\cdot)$ is the logarithmic loss function.

\textbf{HED-SC channel} The HED-SC channel performs the edge detection on the pixel-wise prediction basis. First of all, the lower layer representation of most neural network lacks of semantic meaning due to the gradients vanishing/exploding problem during back-propagation. Deep supervised networks solve this exact problem by adding loss layers in lower structure of the network. In our edge detection model, prior to each pooling layer, feature maps are executed with convolution operation with the kernel size of $3\times 3$, yielding five heatmaps in this case. The prediction for each side-output is calculated as follows:
\begin{equation}
P^{(m)}_{e}\left(z^{*(m)}_{j}=1 \mid X;w,w^{(m)}_{e} \right)=\sigma \left(h\left(X,w,w^{(m)}_{e} \right) \right),
\end{equation}
$\sigma(\cdot)$ is the sigmoid function. The loss function for side-output is:
\begin{equation}
L^{(m)}_{e}\left(Z^{*},X,w,w^{(m)}_{e} \right)=\sum_{j=1}^{\left|Z^{*} \right|}{l}_{E} \left(P^{(m)}_{e} \left(z^{*(m)}_{j}=1 \mid X;w,w^{(m)}_{e} \right)\right),
\end{equation}
$l_{E}(\cdot)$ is cross entropy loss function. Meanwhile, these five side-outputs are generated from feature maps with various sizes, in doing so the architecture of the network is naturally multi-scale. Weighted concatenating the five-scale side-outputs together (the weight $w^{(0)}_{b}$ is learnable), the low-, middle- and high-level information is integrated to generate the edge map:
\begin{equation}
P^{(0)}_{e}\left(z^{*(0)}_{j}=1 \mid X;w,{w}_{e} \right)=\sigma \left(\sum_{m=1}^{M} w^{(0)(m)}_{e}\cdot h\left(X,w, w^{(m)}_{e}\right) \right),
\end{equation}
and the loss function is:\begin{equation}
L^{(0)}_{e}\left(Z^{*},X,w,w_{e} \right)=\sum_{j=1}^{\left|Z^{*} \right|}{l}_{E} \left(P^{(0)}_{e} \left(z^{*(0)}_{j}=1 \mid X;w,w_{e} \right)\right),
\end{equation}
Our loss function of this stage can be computed as:\begin{equation}
{L}_{e}\left(Z^{*},X,w,{w}_{e} \right)=\sum_{m=1}^{M}L^{(m)}_{e}\left(Z^{*},X,w,w^{(m)}_{e} \right)+L^{(0)}_{e}\left(Z^{*},X,w,{w}_{e} \right),
\end{equation}
Merging side-outputs and weighted-fuse would optimize the edge detection result \cite{xie15hed}, but our priority is not edge detection thus we consider $P^{(0)}_{e}$ as the final edge prediction.

\textbf{Training}
At the training phase we combine the pixel prediction and edge prediction together and obtain the fine-grained pixelwise prediction $Y^{*}_{f}$ as our final result:\begin{equation}
{P}_{f}\left(y^{*}_{fj}=k \mid {O}_{r},O^{(0)}_{e};{w}_{f} \right)={\mu}_{k}\left(h\left({O}_{r},O^{(0)}_{e},{w}_{f} \right) \right),
\end{equation}
where ${O}_{r}=h\left(X,w,{w}_{r} \right)$ and $O^{(0)}_{e}=\sum_{m=1}^{M}w^{(0)(m)}_{e} \cdot h\left(X,w,w^{(m)}_{e} \right)$ Firstly, it concatenates the output of first component, the pixel prediction, and the second component, the edge information, together. Then we apply a fully convolutional neural network to process the segmented images. This network contains four convolutional layers, two pooling layers, three full connected layers which are achieved by convolution and an up-sampling layer. We still choose the logarithmic loss function:
\begin{equation}
L_{f}\left(Y^{*}_{f},O_{r},O^{(0)}_{e},w_{f}\right) = \sum_{j=1}^{\left|Y^{*}_{f}\right|}l_{log}\left(P_{f}\left(y^{*}_{fj}=y_{j} \mid O_{r},O^{(0)}_{e};w_{f}\right)\right).
\end{equation}

\section{Experiment}
\textbf{Experiment data}
The dataset is provided by MICCAI 2015 Gland Segmentation Challenge Contest \cite{sirinukunwattana2016gland,sirinukunwattana2015stochastic} which consists of 165 labeled H\&E stained colorectal cancer histological images. There are 85 images in the training set and 80 in the test sets (test A has 60 images and test B has 20 images). 

\begin{figure}[h!] 
 \centering
 \includegraphics[width=4in]{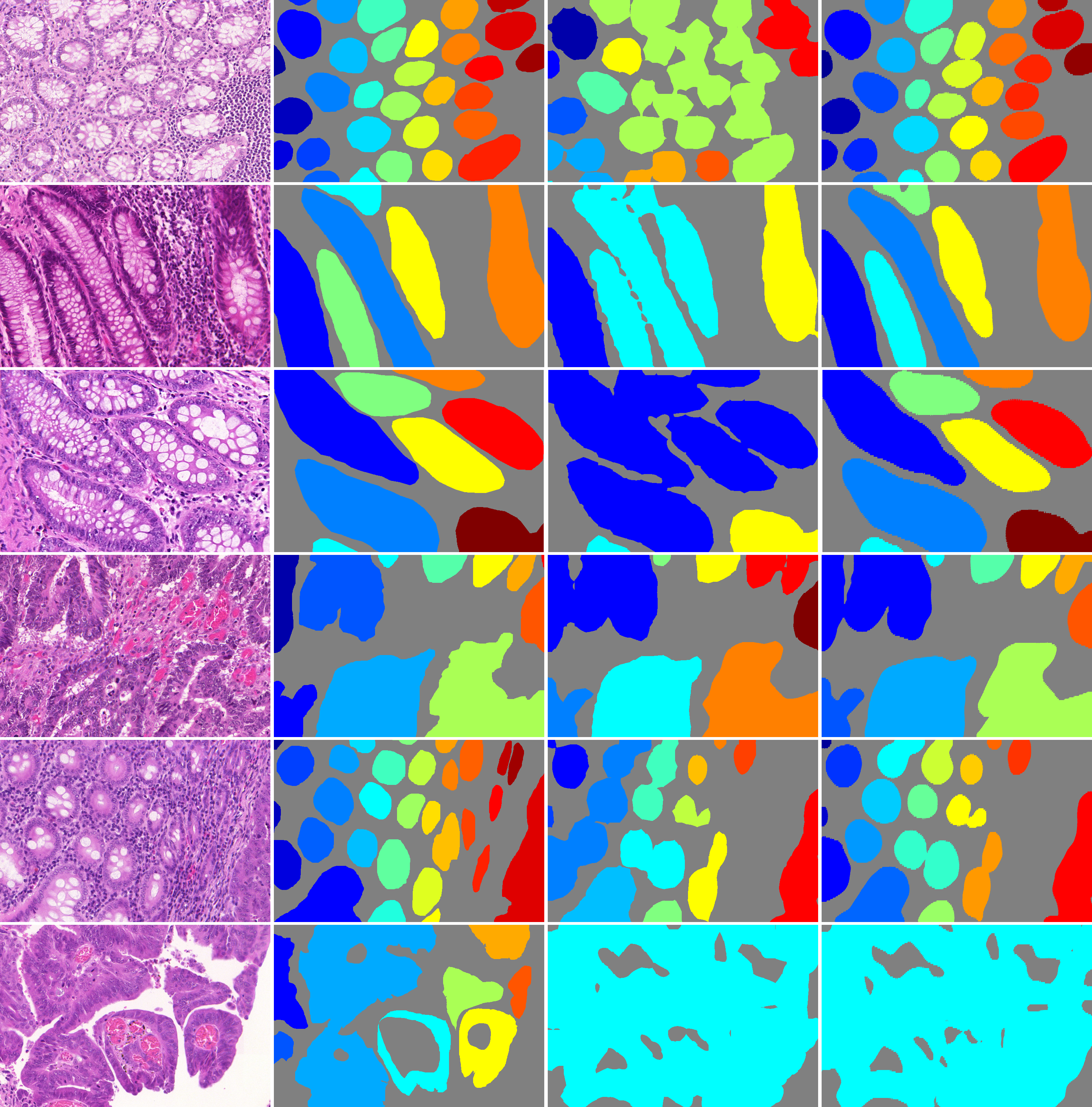} 
 \caption{ From left to right: original image, ground truth, result using FCN, result using DMCS model. Compared to FCN, most of the adjacent glandular structures are separated apart which indicates that our framework accomplishes the instance segmentation goal. However, few glands with small shape or filled with red blood cells escape the detection of our model. The bad performance in the last row is due to the fact that in most samples, the white area are recognized as cytoplasm while in this sample, the white area is the background.}
\end{figure} 

\textbf{Data augmentation}
We first preprocess the data by performing per channel zero mean. To enhance performance and combat overfitting, copious training data are needed to learn the parameters. Given the absence of a large dataset, data augmentation is essential before training. The following lists five methods we deploy in augmentation. Horizontal flipping is used in our given dataset. The insensitivity of orientation in the gland slide enables the rotation operation (0,90,180,270) to training images. Meanwhile, shifting operation is applied to the available training images as well. 

\textbf{Hyperparameters}
We implement our learning network using a deep learning framework CAFFE \cite{jia2014caffe}. Experiments are carried out on K40 GPU and the CUDA edition is 7.0.
During the training phase, backpropagation training is used. The parameters of the framework are as follows: the weight decay is 0.002, the momentum is 0.9, and we choose 10 as the mini-batch size in order to use GPU as efficient as possible. While training the region channel of the network, the learning rate is $10^{-3}$ and the parameters in the framework is initialized by pre-trained FCN32s model \cite{long2015fully}, while the HED-SC channel is trained under the learning rate of $10^{-9}$ and the Xavier initialization is performed. Fusion is learned under the learning rate of $10^{-3}$ and initialized by Xavier initialization. Finally, the whole framework is fine-tuned with the learning rate $10^{-3}$ and the weight of loss of edge is $10^{-6}$.

\textbf{Evaluation}
Three criteria are engaged to evaluate the result of instance segmentation. The summation of six ranking numbers of three criteria on two testing datasets determine the final ranking of each team. The F1 score measures the accuracy of glandular instance segmentation. The true positive is defined as the segmented object which at least 50\% intersects with the ground truth. ObjectDice assesses the performance of segmentation. ObjectHausdorff evaluates the shape similarity between ground truth and segmented object based on object-level Hausdorff distance.
\begin{center}
\begin{table*}[!h]
\resizebox{\textwidth}{!}{ 
\begin{threeparttable}[b]
\begin{tabular}{|c|c|c|c|c|c|c|c|c|c|c|c|c|c|}
  \hline
   \multirow{3}{*}{Method}&
   \multicolumn{4}{c|}{F1 Score}&
   \multicolumn{4}{c|}{ObjectDice}&
   \multicolumn{4}{c|}{ObjectHausdorff}&
   \multirow{3}{*}{Rank Sum}\\
  \cline{2-13}
   &\multicolumn{2}{c|}{Part A}&
   \multicolumn{2}{c|}{Part B}&
   \multicolumn{2}{c|}{Part A}&
   \multicolumn{2}{c|}{Part B}&
   \multicolumn{2}{c|}{Part A}&
   \multicolumn{2}{c|}{Part B}& \\
  \cline{2-13}
  & Score & Rank &  Score & Rank & Score & Rank & Score & Rank & Score & Rank & Score & Rank & \\
  \hline
  FCN & 0.709 & 11 & 0.708 & 5 & 0.748 & 11 & 0.779 & 7 & 129.941 & 12 & 159.639 & 6 & 52 \\
  \hline
  \textbf{\cellcolor[rgb]{.9,.9,.9}Ours} & \cellcolor[rgb]{.9,.9,.9}0.858 & \cellcolor[rgb]{.9,.9,.9}8 &\textbf{\cellcolor[rgb]{.9,.9,.9}0.771} & \textbf{\cellcolor[rgb]{.9,.9,.9}1} & \cellcolor[rgb]{.9,.9,.9}0.888 & \cellcolor[rgb]{.9,.9,.9}2 &\textbf{\cellcolor[rgb]{.9,.9,.9}0.815} & \textbf{\cellcolor[rgb]{.9,.9,.9}1} & \cellcolor[rgb]{.9,.9,.9}54.202 & \cellcolor[rgb]{.9,.9,.9}2 & \textbf{\cellcolor[rgb]{.9,.9,.9}129.930} & \textbf{\cellcolor[rgb]{.9,.9,.9}1} & \cellcolor[rgb]{.9,.9,.9}15\\
  \hline
  CUMedVision2 \cite{chen2016dcan} & \textbf{0.912} & \textbf{1} & 0.716 & 4 & \textbf{0.897} & \textbf{1} & 0.781 & 6 & \textbf{45.418} & \textbf{1} & 160.347 & 8 & 21 \\
  \hline
  ExB1 & 0.891 & 4 & 0.703 & 6 & 0.882  & 5 & 0.786 & 3 & 57.413 & 7 & 145.575 & 2 & 27\\
  \hline
  ExB3 & 0.896 & 2 & 0.719 & 3 & 0.886 & 3 & 0.765 & 8 & 57.350 & 6 & 159.873 & 7 & 29\\
  \hline
  Frerburg2 \cite{ronneberger2015u} & 0.870 & 5 & 0.695 & 7 & 0.876 & 6 & 0.786 & 4 & 57.093 & 4 & 148.463 & 4 & 30\\
  \hline
  CUMedVision1 \cite{chen2016dcan} & 0.868 & 6 & 0.769 & 2 & 0.867 & 9 & 0.800 & 2 & 74.596 & 9 & 153.646 & 5 & 33\\
  \hline
 \end{tabular}
 \caption{Our framework performs outstandingly in datasets provided by MICCAI 2015 Gland Segmentation Challenge Contest and achieves the state-of-the-art result. We rearrange the scores and ranks in this table. Our method outranks FCN and other participants \cite{sirinukunwattana2016gland} based on rank sum.}
\end{threeparttable}}
\end{table*}
\end{center}

\textbf{Result}
Our framework performs well in the dataset provided by MICCAI 2015 Gland Segmentation Challenge and achieves state-of-the-art results (as listed in Table. 1) among all participants \cite{sirinukunwattana2016gland}. We train FCN for 20 epoches in roughly 23h, HED for 20 epoches in 22h, the fusion phase for 10 epoches in 5h and the finetune phase for 40 epoches in 50h. Compared to the result of FCN our framework obtains better score which is a convincing evidence that our work is more effective in solving instance segmentation problem in histological images.

The result of instance segmentation is illustrated in Fig.4. Our method is inspired by FCN and we add the region information to solve the instance segmentation task. Compared to FCN, most of the adjacent glandular structures have been separated apart which indicates that our framework accomplishes the instance segmentation goal. However, glands which are too small and have similar backgrounds (fifth row in Fig.4) are neither detected by FCN nor recognized in the fusion process. Images scattered with red blood cells caused by internal hemorrhage are excluded in training dataset, consequently instance segmentation result (sixth row in Fig.4) is not satisfactory.

\textbf{Discussion}
This framework exploits information from both region and gland channels, of which the region channel accomplishes the segmentation and positioning while the edge channel separates two adjacent gland instances. 

In test A, most of the pathology slide images are the normal ones while test B contains a majority of the images of cancerous tissue which are more complicated in shape and lager in size. Hence, a larger receptive field is required in order to detect cancerous glands. We use 5 pooling layers to enlarge the receptive field but in doing so, the network produces a much smaller heatmap (32 times subsampling of the original image) thus the performance concerning detecting small normal glands gets worse. 

\section{Conclusion}
We propose a new algorithm called deep multichannel side supervision which achieves state-of-the-art results in MICCAI 2015 Gland Segmentation Challenge. The universal framework extracts features of both the edge and region and concatenate them together to generate the result of instance segmentation.

In future work, this algorithm can be utilized in medical images and multichannel learning can be used to improve instance segmentation. 

\section*{Acknowledgement}
This work is supported by Microsoft Research under the eHealth program, the Beijing National Science Foundation in China under Grant 4152033, Beijing Young Talent Project in China, the Fundamental Research Funds for the Central Universities of China under Grant SKLSDE-2015ZX-27 from the State Key Laboratory of Software Development Environment in Beihang University in China. 

\bibliography{gland}   
\bibliographystyle{splncs03}

\end{document}